\documentclass{article}
\usepackage[hyperref]{acl2019}
\usepackage{times}
\usepackage{latexsym}
\usepackage{graphicx}
\usepackage{algorithm2e}
\usepackage{tabularx}
\usepackage{subcaption}
\usepackage{amssymb}
\usepackage{amsfonts}
\usepackage{mathtools}
\usepackage{amsmath}
\usepackage{mathabx}
\usepackage{dblfloatfix}
\usepackage{graphbox}
\usepackage{url}

\aclfinalcopy 
\title{On the Importance of Word Boundaries \\in Character-level Neural Machine Translation\vspace{0.5cm}}

\author{Duygu Ataman$^1$ \\
  University of Trento \\
  Fondazione Bruno Kessler \\
  \texttt{ataman@cl.uzh.ch} \\\And
  Orhan F{\i}rat \\
  Google AI \\
  \texttt{orhanf@google.com} \\\And
  Mattia A. Di Gangi\\
  University of Trento \\
  Fondazione Bruno Kessler \\
  \texttt{digangi@fbk.eu} \\ \AND
  Marcello Federico$^2$ \\
  Amazon AI \\
  \texttt{marcfede@amazon.com} \\\And
  Alexandra Birch\\
  University of Edinburgh \\
  \texttt{a.birch@ed.ac.uk}}

\date{}

\begin{document}
\maketitle
\begin{abstract}
  Neural Machine Translation (NMT) models generally perform translation using a fixed-size lexical vocabulary, which is an important bottleneck on their generalization capability and overall translation quality. The standard approach to overcome this limitation is to segment words into subword units, typically using some external tools with arbitrary heuristics, resulting in vocabulary units not optimized for the translation task. Recent studies have shown that the same approach can be extended to perform NMT directly at the level of characters, which can deliver translation accuracy on-par with subword-based models, on the other hand, this requires relatively deeper networks. In this paper, we propose a more computationally-efficient solution for character-level NMT which implements a hierarchical decoding architecture where translations are subsequently generated at the level of words and characters. We evaluate different methods for open-vocabulary NMT in the machine translation task from English into five languages with distinct morphological typology, and show that the hierarchical decoding model can reach higher translation accuracy than the subword-level NMT model using significantly fewer parameters, while demonstrating better capacity in learning longer-distance contextual and grammatical dependencies than the standard character-level NMT model. \\
\end{abstract}

\hrule
$^1$\small{Work done while the author was a visiting post-graduate research student at the University of Edinburgh.}\\
\indent
$^2$\small{Work done while the author was with Fondazione Bruno Kessler.}\\

\section{Introduction}

Neural Machine Translation (NMT) models are typically trained using a fixed-size lexical vocabulary. In addition to controlling the computational load, this limitation also serves to maintain better distributed representations for the most frequent set of words included in the vocabulary. On the other hand, rare words in the long tail of the lexical distribution are often discarded during translation since they are not found in the vocabulary. The prominent approach to overcome this limitation is to segment words into subword units \cite{sennrich2016neural} and perform translation based on a vocabulary composed of these units. However, subword segmentation methods generally rely on statistical heuristics that lack any linguistic notion. Moreover, they are typically deployed as a pre-processing step before training the NMT model, hence, the predicted set of subword units are essentially not optimized for the translation task.
Recently, \cite{cherry2018revisiting} extended the approach of NMT based on subword units to implement the translation model directly at the level of characters, which could reach comparable performance to the subword-based model, although this would require much larger networks which may be more difficult to train.
The major reason to this requirement may lie behind the fact that treating the characters as individual tokens at the same level and processing the input sequences in linear time increases the difficulty of the learning task, where translation would then be modeled as a mapping between the characters in two languages.
The increased sequence lengths due to processing sentences as sequences of characters also augments the computational cost, and a possible limitation, since sequence models typically have limited capacity in remembering long-distance context.

In many languages, words are the core atomic units of semantic and syntactic structure, and their explicit modeling should be beneficial in learning distributed representations for translation.
There have been early studies in NMT which proposed to perform translation at the level of characters while also regarding the word boundaries in the translation model through a hierarchical decoding procedure, although these approaches were generally deployed through hybrid systems, either as a back-off solution to translate unknown words \cite{luong-hybrid}, or as pre-trained components \cite{DBLP:journals/corr/LingTDB15}.
In this paper, we explore the benefit of achieving character-level NMT by processing sentences at multi-level dynamic time steps defined by the word boundaries, integrating a notion of explicit hierarchy into the decoder.
In our model, all word representations are learned compositionally from character embeddings using bi-directional recurrent neural networks (bi-RNNs) \cite{schuster1997bidirectional}, and decoding is performed by generating each word character by character based on the predicted word representation through a hierarchical beam search algorithm which takes advantage of the hierarchical architecture while generating translations.

We present the results of an extensive evaluation comparing conventional approaches for open-vocabulary NMT in the machine translation task from English into five morphologically-rich languages, where each language belongs to a different language family and has a distinct morphological typology. Our findings show that using the hierarchical decoding approach, the NMT models are able to obtain higher translation accuracy than the subword-based NMT models in many languages while using significantly fewer parameters, where the character-based models implemented with the same computational complexity may still struggle to reach comparable performance. Our analysis also shows that explicit modeling of word boundaries in character-level NMT is advantageous for capturing longer-term contextual dependencies and generalizing to morphological variations in the target language.

\section{Neural Machine Translation}

In this paper, we use recurrent NMT architectures based on the model developed by Bahdanau et al.~\cite{bahdanau2014neural}. 
The model essentially estimates the conditional probability of translating a source sequence $x = (x_1, x_2, \ldots x_m)$ into a target sequence $y = (y_1, y_2, \ldots y_n)$, using the decomposition
\begin{equation}
\label{nmt1}
   p(y|x) = \\
   \prod_{j=1}^n p(y_j|y_{<j},x_m,..,x_1)
\end{equation}

\noindent
where $y_{<j}$ is the target sentence history defined by the sequence $\{y_1...y_{j-1}\}$.

The inputs of the network are {\it one-hot} vectors representing the tokens in the source sentence, which are binary vectors with a single bit set to 1 to identify a specific token in the vocabulary. Each one-hot vector is then mapped to a dense continuous representation, {\it i.e.} an embedding, of the source tokens via a look-up table. The representation of the source sequence is computed using a multi-layer bi-RNN, also referred as the \textit{encoder}, which maps \textit{x} into \textit{m} dense vectors corresponding to the hidden states of the last bi-RNN layer updated in response to the input token embeddings. 

The generation of the translation of the source sentence is called \textit{decoding}, and it is conventionally implemented in an auto-regressive mode, where each token in the target sentence is generated based on an sequential classification procedure defined over the target token vocabulary. In this decoding architecture, a unidirectional recurrent neural network (RNN) predicts the most likely output token $y_{i}$ in the target sequence using an approximate search algorithm based on the previous target token $y_{i-1}$, represented with the embedding of the previous token in the target sequence, the previous decoder hidden state, representing the sequence history, and the current attention context in the source sequence, represented by the \textit{context vector} $c_t$. The latter is a linear combination of the encoder hidden states, whose weights are dynamically computed by a dot product based similarity metric called the \textit{attention model} \cite{luong2015effective}. 

The probability of generating each target word $y_{i}$ is estimated via a softmax function

\begin{equation}
\label{eq:nmt2}
	p(y_i=z_j|x;\theta) = \frac{e^{z_j^T o_i}}{\sum_{k=1}^K e^{z_k^T o_i}} 
\end{equation}

where $z_j$ is the $j^{th}$ one-hot vector of the target vocabulary of size $K$, and $o_i$ is the decoder output vector for the $i^{th}$ target word $y_i$. 
The model is trained by maximizing the log-likelihood of a parallel training set via stochastic gradient-descent \cite{sgd}, where the gradients are computed with the back propagation through time \cite{werbos1990backpropagation} algorithm.

Due to the softmax function in Equation \ref{eq:nmt2}, the size of the target vocabulary plays an important role in defining the computational complexity of the model. In the standard architecture, the embedding matrices account for the vast majority of the network parameters, thus, the amount of embeddings that could be learned and stored efficiently needs to be limited. Moreover, for many words corresponding to the long tail of the lexical distribution, the model fails in learning accurate embeddings, as they are rarely observed in varying context, leading the model vocabulary to typically include the most frequent set of words in the target language. This creates an important bottleneck over the vocabulary coverage of the model, which is especially crucial when translating into low-resource and morphologically-rich languages, which often have a high level of sparsity in the lexical distribution. 

\begin{figure*}[t!]
  \begin{subfigure}[h]{.5\textwidth}
  \centering
    \includegraphics[align=c,scale=0.1]{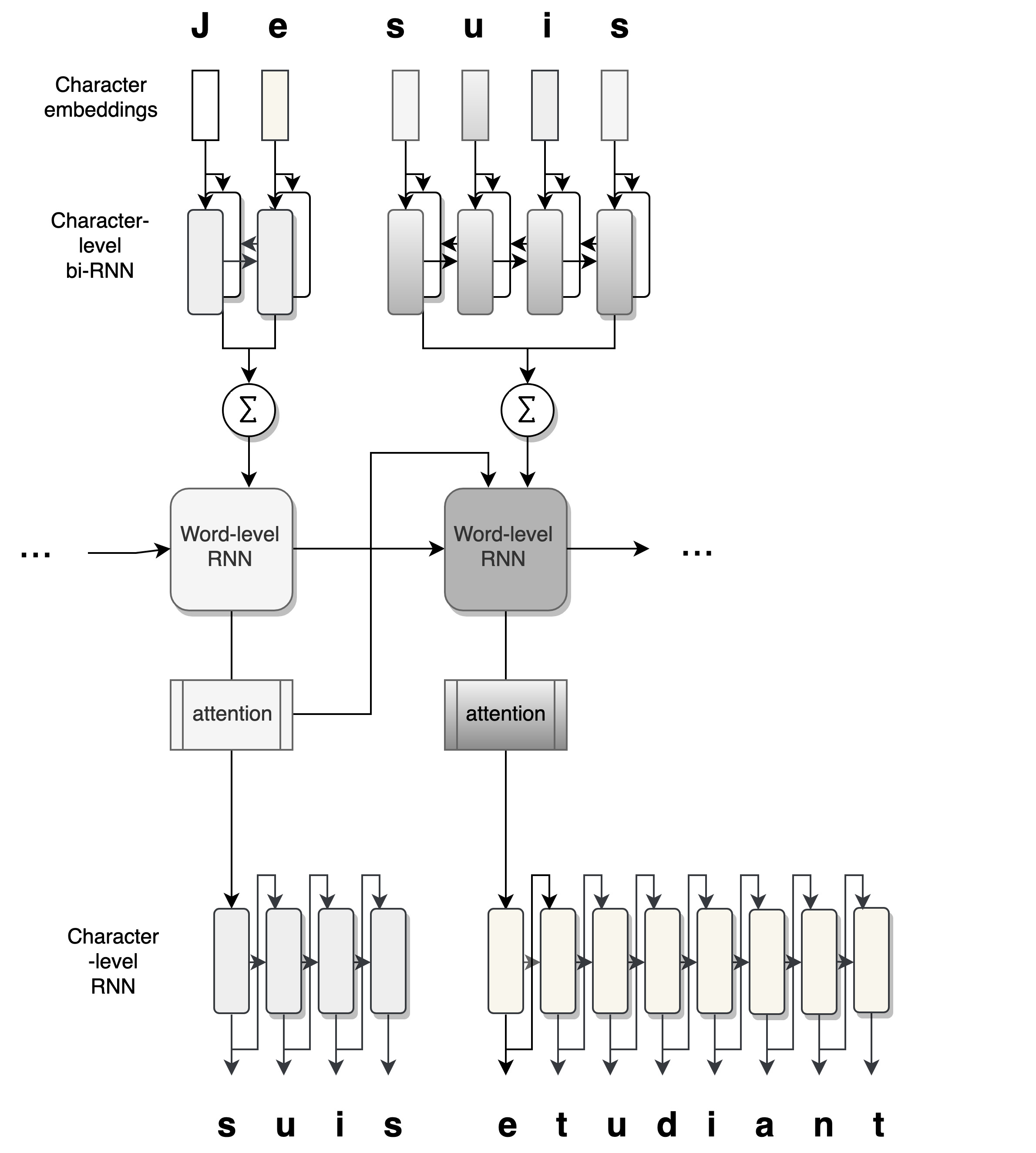}
    \caption{}
    \label{fig:charnmt}
  \end{subfigure}
 \begin{subfigure}[t]{.5\textwidth} 
  \centering
    \includegraphics[align=c,scale=0.1]{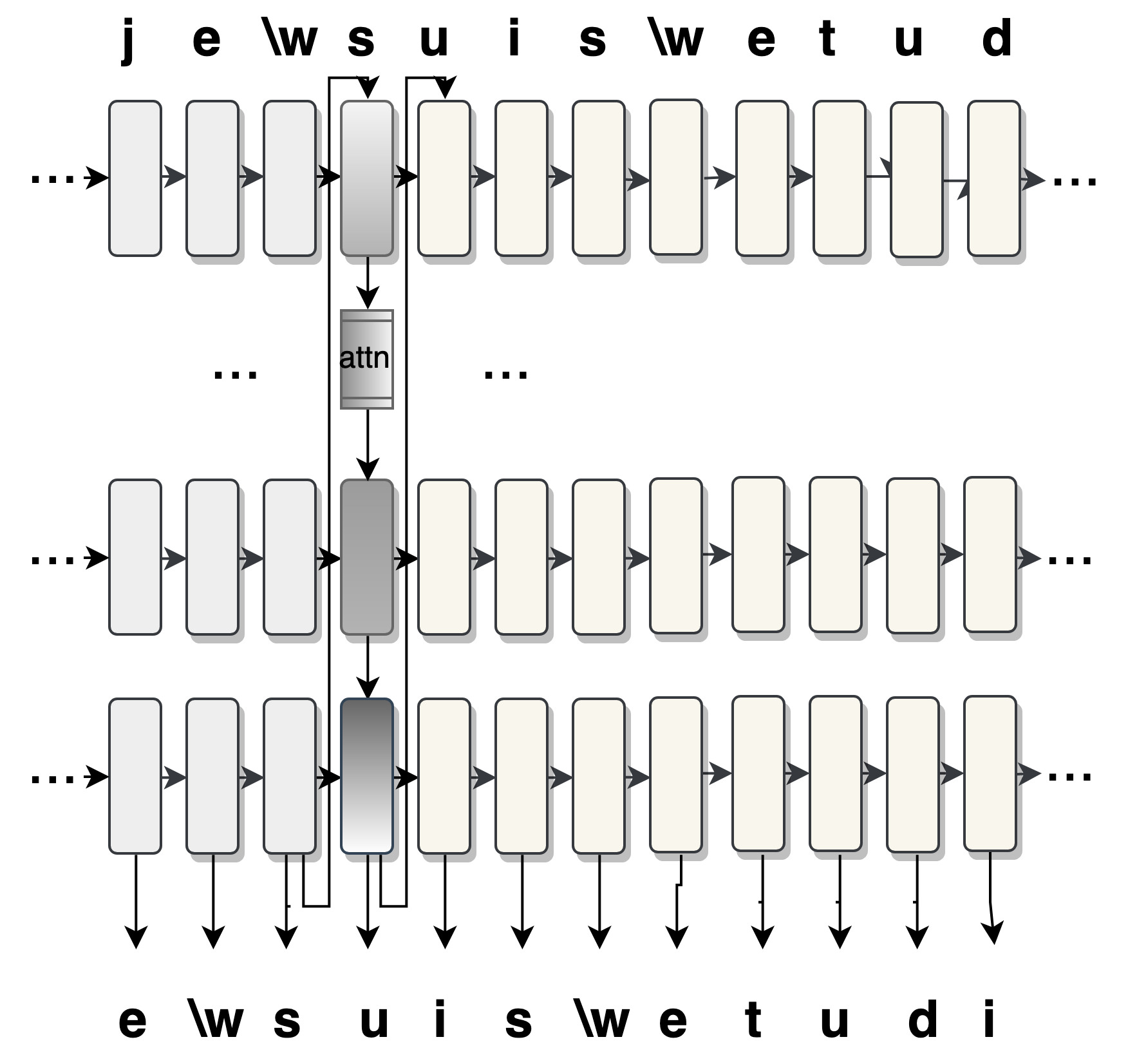}
    \caption{}
    \label{fig:hchar-dec}
  \end{subfigure}
  \caption[Two approaches to character-level NMT]{(a) Hierarchical NMT decoder: input words are encoded as character sequences and the translation is predicted at the level of words. The output words are generated as character sequences. (b) Character-level NMT decoder: the next token in the sentence is predicted by computing the attention weights and the target context repetitively for each character in the sentence. }
  \end{figure*}

The standard approach to overcome this limitation has now become applying a statistical segmentation algorithm on the training corpus which splits words into smaller and more frequent \textit{subword} units, and building the model vocabulary composed of these units. The translation problem is then modeled as a mapping between sequences of subword units in the source and target languages \cite{sennrich2016neural,wu2016google,ataman2017linguistically}. The most popular statistical segmentation method is Byte-Pair Encoding (BPE) \cite{sennrich2016neural}, which finds the optimal description of a corpus vocabulary by iteratively merging the most frequent character sequences. One problem related to the subword-based NMT approach is that segmentation methods are typically implemented as pre-processing steps to NMT, thus, they are not optimized simultaneously with the translation task in an end-to-end fashion. This can lead to morphological errors at different levels, and cause loss of semantic or syntactic information \cite{ataman2017linguistically}, due to the ambiguity in subword embeddings. In fact, recent studies have shown that the same approach can be extended to implement the NMT model directly at the level of characters, which could alleviate potential morphological errors due to subword segmentation. Although character-level NMT models have shown the potential to obtain comparable performance with subword-based NMT models, this would require increasing the computational cost of the model, defined by the network parameters \cite{kreutzer2018learning,cherry2018revisiting}. As given in Figure \ref{fig:charnmt} implementing the NMT decoder directly at the level of characters leads to repetitive passes over the attention mechanism and the RNNs modeling the target language for each character in the sentence. Since the distributed representations of characters are shared among different word and sentence-level context, the translation task requires a network with high capacity to learn this vastly dynamic context.

\section{Hierarchical Decoding}

In this paper, we explore the benefit of integrating a notion of hierarchy into the decoding architecture which could increase the computational efficiency in character-level NMT, following the work of \cite{luong-hybrid}. In this architecture, the input embedding layer of the decoder is augmented with a character-level bi-RNN, which estimates a composition function over the embeddings of the characters in each word in order to compute the distributed representations of target words.

Given a bi-RNN with a forward ($f$) and backward ($b$) layer, the word representation $\mathbf{w}$ of a token of $t$ characters is computed from the hidden states $\mathbf{h}_f^t$ and $\mathbf{h}_b^0$, \textit{i.e.} the final outputs of the forward and backward RNNs, as follows:

\begin{equation}
    \mathbf{w} = \mathbf{W}_f  \mathbf{h}_f^t + \mathbf{W}_b · \mathbf{h}_b^0 + \mathbf{b}
\end{equation}

\noindent
where $\mathbf{W}_f$ and $\mathbf{W}_b$ are weight matrices associated to each RNN and $\mathbf{b}$ is a bias vector. The embeddings of characters and the parameters of the word composition layer are jointly learned while training the NMT model. Since all target word representations are computed compositionally, the hierarchical decoding approach eliminates the necessity of storing word embeddings, significantly reducing the number of parameters.

Each word in the target sentence is predicted by an RNN operating at the level of words, using the compositional target word representations, target sentence history and the context vector computed by the attention mechanism only in the beginning of a new word generation. Instead of classifying the predicted target word in the vocabulary, its distributed representation is fed to a character-level RNN to generate the surface form of the word one character at a time by modeling the probability of observing the $k_{th}$ character of the $j_{th}$ word with length $l$, $p(y_{j,k}|y_{<j},y_{j,<k})$, given the previous words in the sequence and the previous characters in the word. 

The translation probability is then decomposed as:
\begin{align}
   p(y|x) &= \prod_{j=1}^n \prod_{k=1}^{l}  p(y_{j,k}|y_{j,<k},y_{<j},x_{<m})
\end{align}

Similar to \cite{luong-hybrid}, the information necessary to generate the surface form is encoded into the attentional vector $\hat{h}_t$:

\begin{equation}
    \hat{h}_t = \tanh(W[c_t;h_t])
\end{equation}

\noindent
where $h_t$ is the hidden state of the word-level RNN representing the current target context. The attentional vector is used to initialize the character RNN, and after the generation of the first character in the word, character decoding continues in an auto-regressive mode, where the embedding of the each character is fed to the RNN to predict the next character in the word. 
The decoder consecutively iterates over the words and characters in the target sentence, where each RNN is updated at dynamic time steps based on the word boundaries. 

\begin{table*}[t]
\centering
\begin{tabular}{c|ccccc|c}
\hline \hline
    {\bf Model} & \multicolumn{5}{c|}{\bf BLEU} & {\bf Avg. Num.} \\
     & {\bf AR} & {\bf CS} & {\bf DE} & {\bf IT} & {\bf TR} & {\bf Params} \\
      \hline
Subwords & 14.27 &  16.60 &  \bf 24.29 & 26.23 & 8.52 & 22M \\
\hline
Characters & 12.72 & \bf 16.94 & 22.23 & 24.33 & \bf 10.63 & 7.3M \\
\hline
Hierarchical  & \bf 15.55  & 16.79 & 23.91 & \bf 26.64 & 9.74 & 7.3M \\
\hline \hline
\end{tabular}
\caption{Results of the evaluation of models in translating languages with different morphological typology using the IWSLT data sets. The average number of parameters are calculated only for the decoders of the NMT models at a resolution of millions (M). The best scores for each translation direction are in bold font. All improvements over the baselines are statistically significant (p-value~$<$~0.01).}
\label{tab:results1}
\end{table*}

\begin{figure*}[h!]
 \centering
    \includegraphics[scale=0.49]{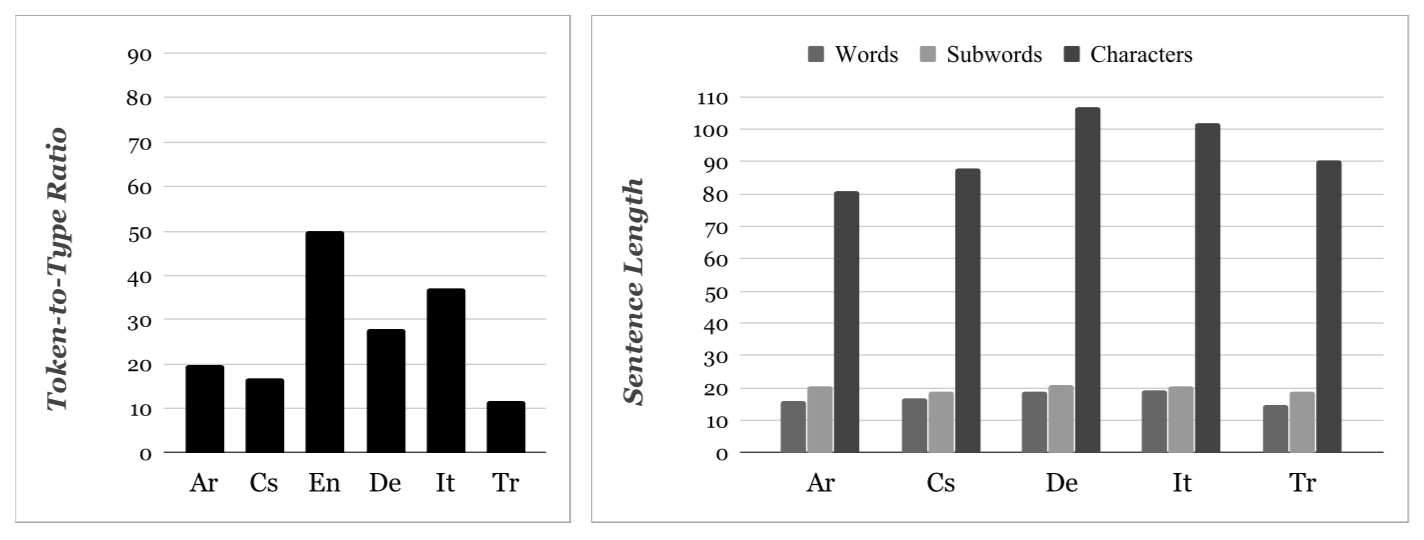}
    \caption{Lexical sparsity and average sentence lengths in different languages.}
    \label{fig:ttr}
  \end{figure*}
  
\section{Hierarchical Beam Search}

\begin{algorithm}[h!]
\resizebox{7cm}{!}{%
    \begin{tabular}{l}
    \hline \hline
    \textbf{function} HierarhicalBeamSearch(Hyp,Best,t) \\
    NewHyp $\leftarrow$ () \\
    \textbf{for all} (seq,score,state) in  Hyp do: \\
    \hspace{2mm} (chars,logpr,$\hat{\textrm{state}}$) $\leftarrow$ CharRNN$_{Fwd}$(tail(seq), state) \\
    \hspace{4mm}\textbf{for all} (c,lp) in  (characters,logpr) do: \\
    \hspace{4mm} hyp=[append(seq,c),score+lp,$\hat{\textrm{state}}$] \\
    \hspace{4mm} \textbf{if} (IsSolution(hyp) \textbf{and}\\
    \hspace{4mm}  hyp.score $>$ Best.score) \\ 
  \hspace{4mm} \textbf{then} Best=hyp \\
     \hspace{4mm} \textbf{else} Push(NewHyp,hyp) \\
    \hspace{2mm} NewHyp $\leftarrow$  Prune(NewHyp,Best)\\
    NewHyp $\leftarrow$ TopB(NewHyp)\\
    NewHyp.state $\leftarrow$ WordRNN$_{Fwd}$(NewHyp)\\
    
    \textbf{if} (NewHyp)\\
    \hspace{2mm} \textbf{return} BeamSearch(NewHyp,Best,t+1)\\
    \textbf{else} \textbf{return} Best\\
     \hline\hline
    \end{tabular}}
    \caption{Hierarchical beam search algorithm. }
    \label{beamsearch}
\end{algorithm}

In order to achieve efficient decoding with the hierarchical NMT decoder, we implement a \textit{hierarchical beam search} algorithm. Similar to the standard algorithm, the beam search starts by predicting the $B$ most likely characters and storing them in a character beam along with their probabilities. The beams are reset each time the generation of a word is complete and the $B$ most likely words are used to update the hidden states of the word-level RNN, which are fed to the character RNN to continue the beam search. When the beam search is complete, the most likely character sequence is generated as the best hypothesis.

\section{Experiments}

We evaluate decoding architectures using different levels of granularity in the vocabulary units and the attention mechanism, including the standard decoding architecture implemented either with subword \cite{sennrich2016neural} or fully character-level \cite{cherry2018revisiting} units, which constitute the baseline approaches, and the hierarchical decoding architecture, by implementing all in Pytorch \cite{paszke2017automatic} within the OpenNMT-py framework \cite{klein2017opennmt}. In order to evaluate how each generative method performs in languages with different morphological typology, we model the machine translation task from English into five languages from different language families and exhibiting distinct morphological typology: Arabic (\textit{templatic}), Czech (\textit{mostly fusional, partially agglutinative}), German (\textit{fusional}), Italian (\textit{fusional}) and Turkish (\textit{agglutinative}). We use the TED Talks corpora \cite{mauro2012wit3} for training the NMT models, which range from 110K to 240K sentences, and the official development and test sets from IWSLT\footnote{The International Workshop on Spoken Language Translation.} \cite{mauro2017overview}. The low-resource settings for the training data allows us to examine the quality of the internal representations learned by each decoder under high data sparseness. 
In order to evaluate how the performance of each method scales with increasing data size, we evaluate the models also by training with a multi-domain training data using the public data sets from WMT\footnote{The Conference on Machine Translation, with shared task organized for news translation.} \cite{bojar2016findings} in the English-to-German direction, followed by an analysis on each model's capability in generalizing to morphological variations in the target language, using the Morpheval \cite{burlot2018wmt} evaluation sets. 

All models are implemented using gated recurrent units (GRU) \cite{cho2014properties} with the same number of parameters. The hierarchical decoding model implements a 3-layer GRU architecture, which is compared with a character-level decoder which also uses a 3-layer stacked GRU architecture. The subword-level decoder has a 2-layer stacked GRU architecture, to account also for the larger number of embedding parameters. The models using the standard architecture have the attention mechanism after the first GRU layer, and have residual connections after the second layer \cite{barone2017deep}. The hierarchical decoder implements the attention mechanism after the second layer and has a residual connection between the first and second layers.
  
The source sides of the data used for training character-level NMT models are segmented using BPE with 16,000 merge rules on the IWSLT data, and 32,000 on WMT. For subword-based models we learn shared merging rules for BPE for 16,000 (in IWSLT) and 32,000 (in WMT) units.
The models use an embedding and hidden unit size of 512 under low-resource (IWSLT) and 1024 under high-resource (WMT) settings, and are trained using the Adam \cite{kinga2015method} optimizer with a learning rate of 0.0003 and decay of 0.5, batch size of 100 and a dropout of 0.2. 
Decoding in all models is performed with a beam size of 5.
The accuracy of each output is measured in terms of the BLEU metric \cite{papineni2002bleu} and the significance of the improvements are measured using bootstrap hypothesis testing \cite{wasserman1989bootstrapping}.

\begin{table*}[t]
\centering
\begin{tabular}{c|c|c|c}
\hline \hline
{\bf Variation} & \textbf{Chars} & \textbf{Subwords} & \textbf{Hier.} \\
\hline
\bf Paradigm contrast features & & & \\
Positive vs. comparative adjective  & \bf 71.4 & 68.4 & 70.1\\
Present vs. future tense & 85.7& \bf 92.0 & 90.6\\
Negation & \bf 97.8 & 97.0 & 94.8\\
Singular vs. plural noun & 88.2 & \bf 88.8 & 88.6\\
Present vs. past tense & 92.0& 93.3& \bf 95.4\\
Compound generation & 60.2& \bf 65.4& 57.8\\
Indicative vs. conditional mode & 86.4& 88.2 & \bf 92.3\\
\bf Average & 83.1 & \bf 84.7 & 84.2\\
\hline
\bf Agreement features & & & \\
Pronoun vs. Nouns (gender) & 96.5 & 97.4& \bf 98.8\\
Pronoun vs. Nouns (number) & 95.4& \bf 96.0& 93.4\\
Pronoun (plural) & 88.6 & \bf 94.3 & 92.2 \\
Pronoun (relative-gender) & 74.2 & 76.4& \bf 78.9\\
Pronoun (relative-number) & 84.2& \bf 90.2 & 87.0\\
Positive vs. superlative adjective & 76.2& 68.2& \bf 80.4\\
Simple vs. coordinated verbs (number) & \bf 96.4& 93.4& 97.2\\
Simple vs. coordinated verbs (person) & 92.3& 92.8& \bf 93.5\\
Simple vs. coordinated verbs (tense) & 82.4 & 86.0&  \bf 90.2\\
\bf Average & 87.4 & 88.3 & \bf 90.17\\
\hline \hline
\end{tabular}
\caption{Results of the evaluation of models in capturing morphological variations in the output using the Morpheval English-German test set. The accuracy is measured with the percentage of correctly captured morphological contrasts. The best scores for each translation direction are in bold font.}
\label{tab:results-morpheval}
\end{table*}

\section{Results}

The results of the experiments given in Table \ref{tab:results1} show that the hierarchical decoder can reach performance comparable to or better than the NMT model based on subword units in all languages while using almost three times less number of parameters. The improvements are especially evident in Arabic and Turkish, languages with the most complex morphology, where the accuracy with the hierarchical decoder is \textbf{1.28} and \textbf{1.22} BLEU points higher, respectively, and comparable in Czech, Italian and German, which represent the fusional languages. In Czech, the hierarchical model outperforms the subword-based model by \textbf{0.19} BLEU and in Italian by \textbf{0.41} BLEU points. The subword-based NMT model achieves the best performance in German, a language that is rich in compounding, where explicit subword segmentation might allow learning better representations for translation units.

The fully character-level NMT model, on the other hand, obtains higher translation accuracy than the hierarchical model in Turkish, with an improvement of \textbf{0.91} BLEU, and in Czech with \textbf{0.15} BLEU points. As can be seen in the statistical characteristics of the training sets illustrated by plotting the token-to-type ratios in each language (Figure \ref{fig:ttr}), these two directions constitute the most sparse settings, where Turkish has the highest amount of sparsity in the benchmark, followed by Czech, and the improvements seem to be proportional to the amount of sparsity in the language. This suggests that in case of high lexical sparsity, learning to translate based on representations of characters might aid in reducing contextual sparsity, allowing to learn better distributed representations.
As the training data size increases, one would expect the likelihood of observing rare words to decrease, especially in languages with low morphological complexity, along with the significance of representing rare and unseen words \cite{cherry2018revisiting}. 
Our results support this hypothesis, where decreasing lexical sparsity, either in the form of the training data size, or the morphological complexity of the target language, eliminates the advantage of character-level translation. In Arabic and Italian, where the training data is almost twice as large as the other languages, using the hierarchical model provides improvements of \textbf{2.83} and \textbf{2.31} BLEU points over the character-level NMT model. In German, the fully character-level NMT model still achieves the lowest accuracy, with \textbf{2.06} BLEU points below the subword-based model. This might be due to the increased level of contextual ambiguity leading to difficulty in learning reliable character embeddings when the model is trained over larger corpora. Another factor which might affect the lower performance of character-level models is the average sentence lengths, which are much longer compared to the sentence lengths resulting from with subword segmentation (Figure \ref{fig:ttr}).

\begin{table}[t]
\centering
\begin{tabular}{c|c}
\hline \hline
{\bf Model} & \textbf{newstest15}  \\
\hline
Subwords & \bf 22.71\\
\hline
Characters & 20.34 \\
\hline
Hierarchical & 22.19 \\
\hline \hline
\end{tabular}
\caption{Experiment results in the English-to-German direction with WMT data sets. Translation accuracy is measured with BLEU. Best scores are in bold font.}
\label{tab:results2}
\end{table}

\begin{table*}[t!]
\begin{small}
\begin{center}
\begin{tabular}{c|c}
\hline \hline
\textbf{Input} & when a friend of mine told me that I needed to \\
& see this great video about a guy protesting bicycle fines \\
& in New York City, I admit I wasn't very interested. \\
\hline
\textbf{Output} & \bf bir arkada\c{s}{\i}m New York'ta bisiklet protestosunu \\
\textbf{\it Subword-based} & protesto etmek \bf i\c{c}in bu filmi izlemeye \\
\textbf{\it Decoder}& \bf ihtiyac{\i}m oldu\u{g}unu s\"{o}ylemi\c{s}ti.\\
\hline
\textbf{Output} & {\bf bana bir arkada\c{s}{\i}m} bana \bf New York'ta bir adam ile ilgili \\
\textbf{\it Character-based} & bir adam hakk{\i}nda {\bf g\"{o}rmem gereken bir} adam hakk{\i}nda\\
\textbf{\it Decoder} & g\"{o}rmem gerekti\u{g}ini \bf s\"{o}yledi. \\
\hline
\textbf{Output} & \bf bir arkada\c{s}{\i}m New York'ta bisiklet yapmaya \\
\textbf{\it Hierarchical} & \bf ihtiyac{\i}m oldu\u{g}unu s\"{o}yledi\u{g}i zaman, \\
{\bf \it Decoder} & {\bf kabul ettim}. \\
\hline
\textbf{Reference} & bir arkada\c{s}{\i}m New York \c{s}ehrindeki bisiklet cezalar{\i}n{\i} protesto \\
& eden bir adam{\i}n bu harika videosunu izlemem gerekti\u{g}ini \\
& s\"{o}yledi\u{g}inde, kabul etmeliyim ki \c{c}ok da ilgilenmemi\c{s}tim.\\
\hline \hline
\end{tabular}
\caption{Example translations with different approaches in Turkish}
\label{tab:chp8examples2}
\end{center}
\end{small}
\end{table*}

In the experiments conducted in the English-to-German translation direction, the results of which are given in Table \ref{tab:results2}, accuracy obtained with the hierarchical and subword-based NMT decoders significantly increase with the extension of the training data, where the subword-based model obtains the best accuracy, followed by the hierarchical model, and the character-level NMT model obtains significantly lower accuracy compared to both approaches. Studies have shown that character-level NMT models could potentially reach the same performance with the subword-based NMT models \cite{cherry2018revisiting}, although this might require increasing the capacity of the network. 
On the other hand, the consistency in the accuracy obtained using the hierarchical decoding model from low to mid resource settings suggests that explicit modeling of word boundaries aids in achieving a more efficient solution to character-level translation. 

Since solely relying on BLEU scores may not be sufficient in understanding the generative properties of different NMT models, we perform an additional evaluation in order to assess the capacity of models in learning syntactic or morphological dependencies using the Morpheval test suites, which consist of sentence pairs that differ by one morphological contrast, and each output accuracy is measured in terms of the percentage of translations that could convey the morphological contrast in the target language. Table \ref{tab:results-morpheval} lists the performance of different NMT models implementing decoding at the level of subwords, characters, or hierarchical word-character units in capturing variances in each individual morphological paradigm and preserving the agreement between inflected words and their dependent lexical items. The results of our analysis support the benefit of using BPE in German as an open-vocabulary NMT solution, where the subword-based model obtains the highest accuracy in most of the morphological paradigm generation tasks. The character-level model shows to be promising in capturing few morphological features better than the former, such as negation or comparative adjectives, and in capturing agreement features, the hierarchical decoding model generally performs better than the subword-based model. The dominance of subword-based models could be due to the high level of compounding in German where segmentation is possibly beneficial in splitting compound words and aiding better syntactic modeling in some cases. These results generally suggest the importance of processing the sentence context at the word level in order to induce a better notion of syntax during generation. 

In order to better illustrate the differences in the outputs of each NMT model, we also present some sample translations in Table \ref{tab:chp8examples2}, obtained by translating English into Turkish using the NMT models trained on the TED Talks corpus. The input sentences are selected such that they are sufficiently long so that one can see the ability of each model in capturing long-distance dependencies in context. The input sentence is from a typical conversation, which requires remembering a long context with many references. We highlight the words in each output that is generated for the first time. Most of the models fail to generate a complete translation, starting to forget the sentence history after the generation of a few words, indicated by the start of generation of repetitions of the previously generated words. The character-level decoder seems to have the shortest memory span, followed by the subword-based decoder, which completely omits the second half of the sentence. Despite omitting the translations of the last four words in the input and some lexical errors, the hierarchical decoder is the only model which can generate a meaningful and grammatically-correct sentence, suggesting that modeling translation based on a context defined at the lexical level might help to learn better grammatical and contextual dependencies, and remembering longer history.

Although current methodology in NMT allows more efficient processing by implementing feed-forward architectures \cite{vaswani2017attention}, our approach can conceptually be applied within these frameworks. In this paper, we limit the evaluation to recurrent architectures for comparison to previous work, including \cite{luong-hybrid}, \cite{sennrich2016neural} and \cite{cherry2018revisiting}, and leave implementation of hierarchical decoding with feed-forward architectures to future work.

\section{Conclusion}
In this paper, we explored the idea of performing the decoding procedure in NMT in a multi-dimensional search space defined by word and character level units via a hierarchical decoding structure and beam search algorithm. Our model obtained comparable to better performance than conventional open-vocabulary NMT solutions, including subword and character-level NMT methods, in many languages while using a significantly smaller number of parameters, showing promising application under high-resource settings. Our software is available for public usage\footnote{https://github.com/d-ataman/Char-NMT}.

\section{Acknowledgments}

This project received funding from the European Union’s Horizon 2020 research and innovation programme under grant agreements 825299 (GoURMET) and 688139 (SUMMA).

\clearpage

\bibliography{acl2019}
\bibliographystyle{acl_natbib}


\end{document}